\documentclass[runningheads]{llncs}
\usepackage[T1]{fontenc}
\usepackage{graphicx}
\usepackage{bm}
\usepackage{bbm}
\usepackage{amssymb}
\usepackage{amsmath}
\usepackage{makecell}
\usepackage{multirow}
\usepackage{textcomp}
\usepackage{textcomp}
\RequirePackage{color}
\usepackage{amssymb}
\usepackage{pifont}
\newcommand{\cmark}{\ding{51}}%
\newcommand{\xmark}{\ding{55}}%
\newcommand{\plus}{\raisebox{.4\height}{\scalebox{.6}{+}}}

\usepackage{marvosym}
\begin{document}
\title{Partially Supervised Unpaired Multi-Modal Learning for Label-Efficient Medical Image Segmentation}
%
%
\author{Lei Zhu\inst{1} \and
Yanyu Xu\inst{2} \and
Huazhu Fu\inst{1} \and
Xinxing Xu\inst{1\text{\Letter}} \and
Rick Siow Mong Goh\inst{1}\and
Yong Liu\inst{1}}

\authorrunning{L. Zhu et al.}
%
\institute{
Institute of High Performance Computing (IHPC), Agency for Science, Technology and Research (A*STAR), 1 Fusionopolis Way, \#16-16 Connexis, Singapore 138632, Republic of Singapore \and
The Joint SDU-NTU Centre for Artificial Intelligence Research (C-FAIR), Shandong University, Shandong, China \\ \email{xuxinx@ihpc.a-star.edu.sg}
}

%
\maketitle              
\begin{abstract}
Unpaired Multi-Modal Learning (UMML) which leverages unpaired multi-modal data to boost model performance on each individual modality has attracted a lot of research interests in medical image analysis. However, existing UMML methods require multi-modal datasets to be fully labeled, which incurs tremendous annotation cost. In this paper, we investigate the use of partially labeled data for label-efficient unpaired multi-modal learning, which can reduce the annotation cost by up to one half. We term the new learning paradigm as Partially Supervised Unpaired Multi-Modal Learning (PSUMML) and propose a novel Decomposed partial class adaptation with snapshot Ensembled Self-Training (DEST) framework for it. Specifically, our framework consists of a compact segmentation network with modality specific normalization layers for learning with partially labeled unpaired multi-modal data. The key challenge in PSUMML lies in the complex partial class distribution discrepancy due to partial class annotation, which hinders effective knowledge transfer across modalities. We theoretically analyze this phenomenon with a decomposition theorem and propose a decomposed partial class adaptation technique to precisely align the partially labeled classes across modalities to reduce the distribution discrepancy. We further propose a snapshot ensembled self-training technique to leverage the valuable snapshot models during training to assign pseudo-labels to partially labeled pixels for self-training to boost model performance. We perform extensive experiments under different scenarios of PSUMML for two medical image segmentation tasks, namely cardiac substructure segmentation and abdominal multi-organ segmentation. Our framework outperforms existing methods significantly.
\end{abstract}

\section{Introduction}
Unpaired Multi-Modal Learning (UMML)~\cite{valindria2018multi,dou2020unpaired,yang2023toward} studies the multi-modal learning task with unpaired multi-modal data, which aims to boost model performance on each individual modality and has attracted a lot of research interests in medical image segmentation. Valindria et al.~\cite{valindria2018multi} are the first to investigate different dual stream Convolutional Neural Network (CNN) architectures for unpaired MRI/CT multi-organ segmentation, where they find an "X"-shaped architecture gives the best performance. Dou et al.~\cite{dou2020unpaired} propose a compact "Chilopod"-shaped network with a knowledge distillation loss to learn with unpaired MRI/CT data and achieve state-of-the-art performance on cardiac substructure and multi-organ segmentation. More recently, Yang et al.~\cite{yang2023toward} propose to learn the structured semantic consistency across modalities for unpaired multi-modal learning. While existing UMML methods can effectively exploit the shared cross-modality information to improve model performance for each individual modality, they all require the unpaired multi-modal data to be fully labeled, which incurs tremendous annotation cost. Partially supervised medical image segmentation~\cite{dmitriev2019learning,huang2020multi,shi2021marginal,zhang2022deep} is a well-studied research area in medical domain, where it aims to leverage multiple partially labeled datasets to train a model to predict classes in the joint label set. Konstaintin et al.~\cite{dmitriev2019learning} propose a conditional CNN network which conditions on class information for multi-class segmentation by training with single-class datasets. Huang et al.~\cite{huang2020multi} proposes a co-training framework with weighted-averaged models. Shi et al.~\cite{shi2021marginal} propose marginal and exclusive loss for learning with partially labeled multi-organ datasets. More recently, Zhang et al.~\cite{zhang2022deep} propose deep compatible learning for partially supervised medical image segmentation. However, existing partially supervised segmentation studies have mainly focused on learning with partially labeled datasets from the same modality and fail to handle the complex partial class distribution discrepancy across multiple modalities.

\begin{figure}[t]
\begin{center}
\includegraphics[width=0.75\linewidth]{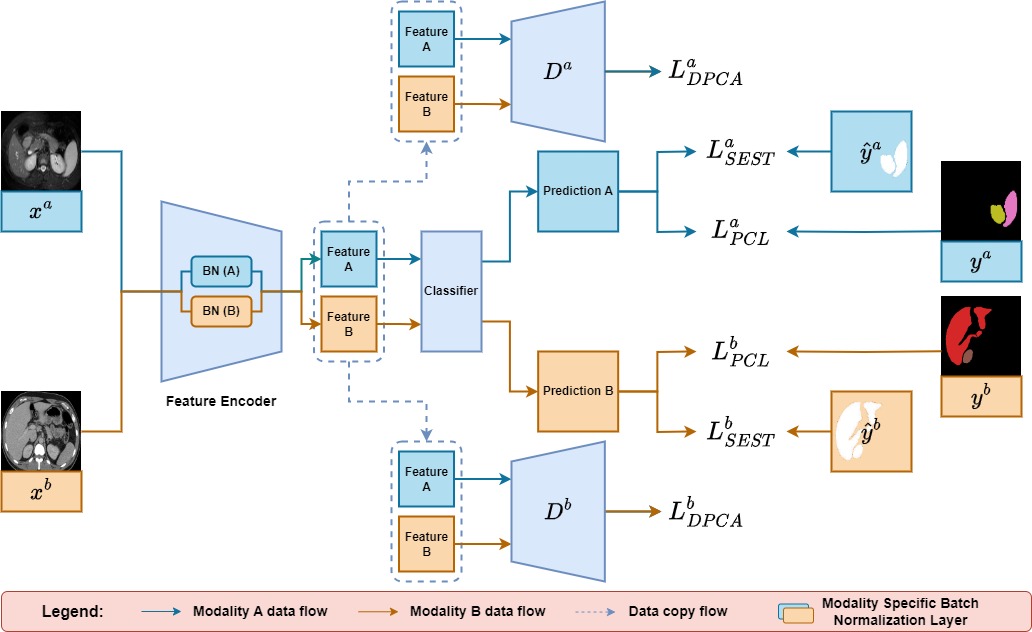}
\end{center}
   \caption{Architecture and dataflow of our proposed DEST framework. DEST consists of one compact segmentation network with modality specific normalization layer and two class conditional domain discriminators with decomposed partial class adaptation loss to reduce partial class distribution discrepancies across modalities. We further propose snapshot ensembled pseudo-labels $\hat{y}^s$ and $\hat{y}^t$ for self-training to boost performance.}
\label{fig:architecture}
\end{figure}

In this paper, we investigate the use of partially labeled data for label-efficient unpaired multi-modal learning. To the best of our knowledge, we are the first to study partially supervised medical image segmentation with unpaired multi-modal datasets. We term the new learning paradigm as Partially Supervised Unpaired Multi-Modal Learning (PSUMML). We propose a novel Decomposed partial class adaptation with snapshot Ensembled Self-Training (DEST) framework for the task. Our framework consists of a compact segmentation network with modality specific normalization layers for learning with partially labeled unpaired multi-modal data. The key challenge in PSUMML lies in the complex partial class distribution discrepancies across modalities due to partial class annotation, which hinders effective knowledge transfer across modalities. We theoretically analyze this phenomenon with a decomposition theorem and propose a decomposed partial class adaptation technique to tackle the partial class distribution discrepancy problem via introducing two class conditional domain discriminators into our framework to precisely align the partially labeled classes across modalities. The segmentation network and the two class conditional domain discriminators play a three-player min-max game at the feature level to reduce the partial class distribution discrepancy, which effectively transfers the labeled class knowledge across modalities where annotation is partially labeled. We further propose a snapshot ensembled self-training technique to leverage the valuable snapshot models during training to assign pseudo-labels to partially labeled pixels for self-training to boost model performance.

In summary, we have made the following contributions in this paper: \textbf{(1).} we introduce a new learning paradigm termed Partially Supervised Unpaired Multi-Modal Learning for label-efficient medical image segmentation, which is of great practical value in medical domain; \textbf{(2).} we provide theoretical analysis on PSUMML and propose a principally designed decomposed partial class adaptation loss function to tackle the complex partial class distribution discrepancy problem in PSUMML; \textbf{(3).} we further propose a snapshot ensembled self-training technique to assign pseudo-labels to partially labeled pixels for self-training to boost model performance; \textbf{(4).} we perform extensive experiments to evaluate our framework under different scenarios of PSUMML on two medical image segmentation tasks. Our framework outperforms existing methods significantly.

\section{Methodology}
We formally define the Partially Supervised Unpaired Multi-Modal Learning (PSUMML) problem as follows: denote $\mathbb{D}^a=\{(x_{i}^a,y_{i}^a)\}^{N^a}_{i=1}$ and $\mathbb{D}^b=\{(x_{i}^b, y_{i}^b)\}^{N^b}_{i=1}$ as two unpaired labeled datasets in modality $a$ and $b$ respectively, where $y_{i}^a$ is the segmentation mask for $x_{i}^a$, similarly for $y_{i}^b$ and $x_{i}^b$. Let $L^a=\{l_0, l^a_1, ..., l^a_{M^a}\}$ be the label set of $\mathbb{D}^a$ and $L^b=\{l_0, l^b_1, ..., l^b_{M^b}\}$ be the label set of $\mathbb{D}^b$, where there are $M^a$ organs (or substructures) which are labeled in $\mathbb{D}^a$ and $M^b$ organs (or substructures) which are labeled in $\mathbb{D}^b$. The unlabeled classes in each modality is all partially labeled as $l_0$, which denotes the "background" class. We assume $L^a \neq L^b$. Let $L=L^a \cup L^b$ be the full label set from the two modalities with the number of classes equals to $M$, namely $M=|L|$, the goal is to learn a model to segment all the classes in $L$ accurately for data from both modalities. 

Note we can further categorize PSUMML into three learning scenarios: (1). $L^a \cap L^b = \varnothing$, where label sets from two modalities are disjoint; (2). $L^a \cap L^b \neq \varnothing \land L^a \not\subset L^b \land L^b \not\subset L^a$, where label sets from two modalities intersect with each other, but neither is a proper superset of the other; (3). $L^a \subset L^b \lor L^b \subset L^a$, where the label set from one modality is a proper superset of the other. 
We present an overview of our proposed Decomposed partial class adaptation with snapshot Ensembled Self-Training (DEST) framework in Fig.~\ref{fig:architecture}. Our framework seamlessly handles all three learning scenarios.
\subsection{Segmentation}
Our framework consists of a segmentation network $S(\cdot):\mathbb{R}^{W\times H\times I}\rightarrow \mathbb{R}^{W\times H\times M}$, which is composed with a feature encoder $F(\cdot)$ and a classifier $C(\cdot)$, where $S=C\circ F$.
Multi-modal learning models usually contain modality shared and specific network layers to model multi-modal information~\cite{valindria2018multi}. Similarly, we utilize modality specific batch normalization layers~\cite{Ioffe2015BatchNA} to model the modality specific information and share the rest network layers to model the shared multi-modal information following~\cite{dou2020unpaired}. As annotations are partially labeled, we extend the marginal loss from~\cite{shi2021marginal} for partial class learning with partially labeled unpaired multi-modal data. Specifically, the background class in each partially labeled multi-modal dataset is a merged class of all partially labeled organs (or substructures) and the true background class, thus the probability of the merged class is a marginal probability. Denote $L^{a'}=L\setminus L^a$ as partially labeled classes in modality $a$. We define an operation $A^{L^{a'}}(\cdot):\mathbb{R}^{W\times H \times M}\rightarrow \mathbb{R}^{W\times H \times (M^a+1)}$, which takes the output prediction mask from $S$ and adds the prediction probabilities in $L^{a'}$ with the true background class. Similarly, we define $L^{b'}$ and $A^{L^{b'}}(\cdot)$. The partial class learning loss for partially labeled unpaired multi-modal data is defined as follows:
\begin{align}
\begin{split}
\mathcal{L}_{PCL}=\mathcal{L}_{PCL}^a+\mathcal{L}_{PCL}^b&=\mathbb{E} [H(y^a,A^{L^{a'}}(S(\bm{x}^a))) + Dice(y^a, A^{L^{a'}}(S(\bm{x}^a)))]\\
&+\mathbb{E} [H(y^b,A^{L^{b'}}(S(\bm{x}^b))) + Dice(y^b, A^{L^{b'}}(S(\bm{x}^b)))],\label{plcpa_eqn:1}
\end{split}
\end{align}
where $H(\cdot)$ is the pixel-wise cross-entropy loss and $Dice(\cdot)$ is the dice loss~\cite{milletari2016v}. 

\subsection{Decomposed Partial Class Adaptation}
We theoretically analyze the PSUMML problem with the following theorem:
\begin{theorem} [Decomposition Theorem on PSUMML] \label{plcpa_theorem:2} Following the problem definition, suppose $L^{b'}$ and $L^{a'}$ are both non-empty set. Denote $\widehat{\mathbb{D}}^a$ as $\mathbb{D}^a$ with label set $L^{b'}\cup \{l_0\}$ and $\widehat{\mathbb{D}}^b$ as $\mathbb{D}^b$ with label set $L^{a'}\cup \{l_0\}$. Let $\mathcal{F}\subseteq \mathbb{R}^{\mathcal{X}\times\mathcal{Y}}$ be the hypothesis set of scoring functions with $\mathcal{Y}=L$, then for any $\delta > 0$, with probability $1-3\delta$, we have the following uniform generalization bound for all scoring functions $f$:
\begin{align}
\label{plcpa_eqn:7}
\begin{split}
\epsilon_{\mathcal{D}^a}&(f)+\epsilon_{\mathcal{D}^b}(f)\leq\epsilon_{\mathbb{D}^a}^{(\rho)}(f)+\epsilon_{\mathbb{D}^b}^{(\rho)}(f)+\epsilon_{\widehat{\mathbb{D}}^a}^{(\rho)}(f)+\epsilon_{\widehat{\mathbb{D}}^b}^{(\rho)}(f)\\
&+d_{f,\mathcal{F}}^{(\rho)}(\widehat{\mathbb{D}}^a, \mathbb{D}^b)+d_{f,\mathcal{F}}^{(\rho)}(\widehat{\mathbb{D}}^b,\mathbb{D}^a)+\lambda_1+\lambda_2 \\
&+\frac{4k^2}{\rho}\mathfrak{R}_{N^a,\mathcal{D}^a}(\Pi_1\mathcal{F})+\frac{4k}{\rho}\mathfrak{R}_{N^a,\mathcal{D}^a}(\Pi_{\mathcal{H}}\mathcal{F})+4\sqrt{\frac{log{\frac{2}{\delta}}}{2N^a}}\\
&+\frac{4k^2}{\rho}\mathfrak{R}_{N^b,\mathcal{D}^b}(\Pi_1\mathcal{F})+\frac{4k}{\rho}\mathfrak{R}_{N^b,\mathcal{D}^b}(\Pi_{\mathcal{H}}\mathcal{F})+4\sqrt{\frac{log{\frac{2}{\delta}}}{2N^b}},\\
\end{split}
\end{align}
where $\epsilon_{\cdot}(\cdot)$ measures the expected error of a hypothesis on a data distribution, $\epsilon_{\cdot}^{(\rho)}(\cdot)$ measures the empirical margin error of a hypothesis on a dataset, $d_{f,\mathcal{F}}^{(\rho)}(\cdot, \cdot)$ is the empirical estimation of the Margin Disparity Discrepancy between two datasets, 
$\lambda_1=\min_{f \in \mathcal{H}} \epsilon_{\widehat{\mathcal{D}}^a}^{(\rho)}(f)+\epsilon_{\mathcal{D}^b}^{(\rho)}(f)$ and $\lambda_2=\min_{f \in \mathcal{H}} \epsilon_{\widehat{\mathcal{D}}^b}^{(\rho)}(f)+\epsilon_{\mathcal{D}^a}^{(\rho)}(f)$ are the ideal joint margin errors, $\mathfrak{R}(\cdot)$ measures the Rademacher Complexity of a function space, $\Pi_1\mathcal{F}=\{x\rightarrow f(x,y)|y\in\mathcal{Y},f\in\mathcal{F}\}$, $\Pi_{\mathcal{H}}\mathcal{F}=\{x\rightarrow f(x,h(x)|h\in\mathcal{H},f\in\mathcal{F}\}$ and $\rho$ is the margin.
\end{theorem}
The theorem decomposes the multi-modal error of a hypothesis on the full label set and upper bounds it with: \textbf{(1).} the empirical margin error on the partial label sets in each modality; \textbf{(2).} the partial class distribution discrepancy in each modality; \textbf{(3).} the ideal joint errors; \textbf{(4).} the complexity terms. Note the ideal joint errors are assumed to be small and the complexity terms are not optimizable. The partial class learning loss in the previous section minimizes the empirical margin errors on the partial label sets. However, the theorem further indicates that it is necessary to minimize the two partial class distribution discrepancy terms, namely $d_{f,\mathcal{F}}^{(\rho)}(\widehat{\mathbb{D}}^a, \mathbb{D}^b)$ and $d_{f,\mathcal{F}}^{(\rho)}(\widehat{\mathbb{D}}^b,\mathbb{D}^a)$. 

To this end, we propose a decomposed partial class adaptation loss function to precisely minimize the two partial class distribution discrepancy terms in the upper bound. Suppose $L^{a'}$ is not empty, we define a domain discriminator $D^a(\cdot):\mathbb{R}^{W'\times H'\times I'}\rightarrow \mathbb{R}^{W\times H\times 2(M-M^a)}$, which takes the image feature as input and outputs a per-pixel class-aware domain prediction. Similarly as the partial class learning loss, we merge the labeled class prediction from the segmentation network and concatenate it with $\mathbf{0}$ masks as ground truth to train the domain discriminator to distinguish the partially labeled classes in modality $a$ and its labeled counterparts across modality. The adversarial class-conditional loss for domain discriminator $D^a$ is defined as follows:
\begin{align}
\mathcal{L}_{DPCA_{D^a}}^{a}=
    \begin{cases}
       &0 \qquad\qquad\qquad\qquad\qquad\qquad\qquad\quad\;\;\:\text{if $L^{a'}$ is empty,}\\
        \begin{split}
       &\mathbb{E} [H([A^{L^a}(S(\bm{x}^b));\mathbf{0}], D^a(F(\bm{x}^b))) ]+ \\ &\mathbb{E} [H([\mathbf{0};A^{L^a}(S(\bm{x}^a))], D^a(F(\bm{x}^a)))]
        \end{split} \qquad\quad \text{otherwise.}
    \end{cases} \label{plcpa_eqn:2}
\end{align}
The adversarial class-conditional loss for feature encoder $F$ is defined as follows:
\begin{align}
\mathcal{L}_{DPCA_{F}}^{a}=
    \begin{cases}
       &0 \qquad\qquad\qquad\qquad\qquad\qquad\qquad\quad\: \text{if $L^{a'}$ is empty,}\\
        \begin{split}
       &\mathbb{E} [H([A^{L^a}(S(\bm{x}^a));\mathbf{0}], D^a(F(\bm{x}^a)))]
        \end{split} \qquad\quad \text{otherwise.}
    \end{cases} \label{plcpa_eqn:3}
\end{align}
Note when $L^{a'}$ is empty, there is no need to minimize the partial class distribution discrepancy. For simplicity, we use $\mathcal{L}_{DPCA}^{a}$ to denote both $\mathcal{L}_{DPCA_{D^a}}^{a}$ and $\mathcal{L}_{DPCA_{F}}^{a}$ without ambiguity. Similarly, we define domain discriminator $D^b$ and adversarial class-conditional loss for $D^b$ and $F$ as $\mathcal{L}_{DPCA}^{b}$. The two class conditional domain discriminators and feature encoder play a three-player min-max game on the loss function following the Generative Adversarial Nets (GAN)~\cite{goodfellow2014generative} framework. At the optimal, the feature encoder aligns the partially labeled class distribution in each modality towards their labeled counterparts across modality so that the two partial class distribution discrepancy terms are minimized.
\subsection{Snapshot Ensembled Self-Training}
Decomposed partial class adaptation reduces partial class distribution discrepancies across modalities to transfer knowledge, but the learned features can still be non-discriminative for the partially labeled classes in each modality, which can negatively affect the ideal joint error term in Theorem~\ref{plcpa_theorem:2}. To this end, we propose snapshot ensembled self-training to assign pseudo-labels to partially labeled pixels for self-training. To rectify pseudo-label noise, inspired by the fact that ensemble of neural networks are more accurate than individual neural network~\cite{zhou2012ensemble}, we adopt snapshot ensembling~\cite{huang2017snapshot}, where we take freely available $K$ time-snapshots of segmentation network during training and ensemble their predictions as soft pseudo-labels. Specifically, we take equally spaced $K$ time-snapshots of network $\{S_{t_1}, ..., S_{t_K}\}$ from time step $t_1$ to $t_K$ after some initial time step $t_0$. The initial time step is chosen to ensure that the network has learned enough segmentation knowledge. 
Then, for input image $\bm{x}$ from each modality, we apply $K$ time-snapshot models to give predictions to it and average the prediction results as a soft pseudo-label denoted as $p_{\bm{x}}=\frac{1}{K}\sum_{i=1}^{K}S_{t_i}(\bm{x})$. The ensembled soft pseudo-label serves as a source of supervision for learning with partially labeled pixels. To further rectify the pseudo-label noise, we utilize current model's prediction to modulate the ensembled soft pseudo-label as $\tilde{p}_{\bm{x}}=S(\bm{x})\odot p_{\bm{x}}$, where $\odot$ is element-wise multiplication. We assign pseudo-labels to partially labeled pixel only when its modulated predicted class probability is higher than a given threshold. 
We denote the operation of assigning pseudo-labels as $\hat{y}=\xi(\tilde{p}_{\bm{x}})$. 
The snapshot ensembled self-training loss is defined as follows:
\begin{equation}
\begin{split}
\mathcal{L}_{SEST}=\mathcal{L}_{SEST}^a+\mathcal{L}_{SEST}^b&=\mathbb{E} [H(\hat{y}^a, S(\bm{x}^a))] + \mathbb{E} [H(\hat{y}^b,S(\bm{x}^b))], \label{eqn:3}
\end{split}
\end{equation}
\textbf{Training objective:} The overall objective of our framework is as follows:
\begin{equation}
    \begin{split}
    \mathcal{L}_{all}=&\mathcal{L}_{PCL}+\lambda (\mathcal{L}_{DPCA}^a+\mathcal{L}_{DPCA}^b) +\mathcal{L}_{SEST}, \label{eqn:9}
    \end{split}
\end{equation}
where $\lambda$ is the balancing weight set to be $0.01$ empirically.

\section{Experimental Analysis}
\noindent\textbf{Datasets and Experimental Settings.} We evaluate the effectiveness of our framework on three datasets for cardiac substructure segmentation~\cite{zhuang2016multi} and abdominal multi-organ segmentation~\cite{kavur2020chaos,landman2017}.
The cardiac dataset consists of unpaired MRI and CT images with ground truth on four heart substructures: ascending aorta (AA), left atrium blood cavity (LAC), left ventricle blood cavity (LVC), and myocardium of the left ventricle (MYO).
The abdominal dataset consists of unpaired T2-SPIR MRI and CT images with ground truth on four organs: spleen, right kidney, left kidney and liver. 
Each dataset is randomly split with $80\%$ scans for training and $20\%$ for testing. We focus on scenario (1) of PSUMML, which can help reduce the most annotation cost compared to the other two. But we also study the other two scenarios. We synthesize two tasks for scenario (1) with the fully labeled datasets by converting labels of certain classes into background class: (i). for cardiac dataset, we let MRI data be partially labeled with AA and LVC and CT data be partially labeled with LAC and MYO; (ii). for abdominal multi-organ dataset, we let MRI data be partially labeled with left kidney and spleen and CT data be partially labeled with liver and right kidney. We employ two commonly-used metrics, the Dice coefficient (Dice) and the average symmetric surface distance (ASD) to quantitatively evaluate the segmentation performance. (Please view the Appendix for more data pre-processing and implementation details of our framework.)

\begin{table}[t]
\caption{Performance comparison on cardiac substructure segmentation and abdominal multi-organ segmentation under PSUMML scenario (1) task in Dice score (\%). 
\textcolor{red}{\textbf{Best}} and \textcolor{blue}{\textbf{second best}} results are in red and blue bold respectively.}
\begin{center}
\resizebox{\linewidth}{!}{%
\begin{tabular}{c|p{0.045\linewidth}p{0.045\linewidth}p{0.045\linewidth}p{0.045\linewidth}p{0.045\linewidth}p{0.045\linewidth}p{0.045\linewidth}p{0.045\linewidth}p{0.045\linewidth}p{0.045\linewidth}p{0.045\linewidth}p{0.045\linewidth}p{0.045\linewidth}p{0.045\linewidth}|p{0.045\linewidth}}
\hline
\multirow{2}{*}{Method} &\multicolumn{1}{c|}{Anno.} & \multicolumn{5}{c}{\textbf{Cardiac Segmentation (MRI $/$ CT)}} & \multicolumn{5}{|c}{\textbf{Multi-Organ Segmentation (MRI $/$ CT)}} \\
\cline{3-12}
&\multicolumn{1}{c|}{Cost (\%)}&\multicolumn{1}{c}{AA}&\multicolumn{1}{c}{LAC}&\multicolumn{1}{c}{LVC}&\multicolumn{1}{c}{MYO}&\multicolumn{1}{c}{Mean}&\multicolumn{1}{|c}{Liver}&\multicolumn{1}{c}{R. kidney}&\multicolumn{1}{c}{L. kidney}&\multicolumn{1}{c}{Spleen}&\multicolumn{1}{c}{Mean} \\
\hline
Supervised (partial)&\multicolumn{1}{c|}{50}&\multicolumn{1}{c}{77.5$/$28.4}&\multicolumn{1}{c}{0.9$/$90.1}&\multicolumn{1}{c}{91.4$/$0.1}&\multicolumn{1}{c}{0.0$/$87.4}&\multicolumn{1}{c}{42.5$/$51.5}&\multicolumn{1}{|c}{69.6$/$95.2}&\multicolumn{1}{c}{3.1$/$90.3}&\multicolumn{1}{c}{91.4$/$27.0}&\multicolumn{1}{c}{90.7$/$54.7}&\multicolumn{1}{c}{63.7$/$66.8} \\
\Xhline{4\arrayrulewidth}
ConCNN~\cite{dmitriev2019learning}&\multicolumn{1}{c|}{50}&\multicolumn{1}{c}{80.4$/$0.0}&\multicolumn{1}{c}{0.0$/$89.0}&\multicolumn{1}{c}{85.3$/$0.0}&\multicolumn{1}{c}{0.0$/$8.6}&\multicolumn{1}{c}{41.4$/$24.4}&\multicolumn{1}{|c}{0.0$/$26.4}&\multicolumn{1}{c}{0.0$/$40.8}&\multicolumn{1}{c}{69.7$/$0.0}&\multicolumn{1}{c}{33.7$/$0.0}&\multicolumn{1}{c}{25.8$/$16.8} \\
CoWM~\cite{huang2020multi}&\multicolumn{1}{c|}{50}&\multicolumn{1}{c}{83.1$/$49.3}&\multicolumn{1}{c}{0.9$/$90.2}&\multicolumn{1}{c}{92.9$/$0.1}&\multicolumn{1}{c}{0.0$/$86.4}&\multicolumn{1}{c}{44.2$/$56.5}&\multicolumn{1}{|c}{77.3$/$93.5}&\multicolumn{1}{c}{19.9$/$89.2}&\multicolumn{1}{c}{92.8$/$68.0}&\multicolumn{1}{c}{92.6$/$68.0}&\multicolumn{1}{c}{70.6$/$79.7} \\
Marginal~\cite{shi2021marginal}&\multicolumn{1}{c|}{50}&\multicolumn{1}{c}{82.6$/$77.0}&\multicolumn{1}{c}{11.2$/$90.2}&\multicolumn{1}{c}{90.8$/$36.4}&\multicolumn{1}{c}{3.6$/$86.6}&\multicolumn{1}{c}{47.0$/$72.5}&\multicolumn{1}{|c}{79.9$/$94.4}&\multicolumn{1}{c}{63.9$/$91.7}&\multicolumn{1}{c}{86.2$/$53.3}&\multicolumn{1}{c}{90.5$/$55.3}&\multicolumn{1}{c}{80.1$/$73.7} \\
\hline
X-network~\cite{valindria2018multi}&\multicolumn{1}{c|}{50}&\multicolumn{1}{c}{81.8$/$29.8}&\multicolumn{1}{c}{0.2$/$90.5}&\multicolumn{1}{c}{92.7$/$0.1}&\multicolumn{1}{c}{0.0$/$86.4}&\multicolumn{1}{c}{43.7$/$51.7}&\multicolumn{1}{|c}{71.9$/$95.0}&\multicolumn{1}{c}{3.2$/$91.3}&\multicolumn{1}{c}{92.6$/$41.9}&\multicolumn{1}{c}{92.0$/$46.8}&\multicolumn{1}{c}{64.9$/$68.8}\\
Y-network~\cite{valindria2018multi}&\multicolumn{1}{c|}{50}&\multicolumn{1}{c}{81.5$/$66.0}&\multicolumn{1}{c}{26.6$/$86.9}&\multicolumn{1}{c}{92.4$/$21.0}&\multicolumn{1}{c}{8.7$/$86.2}&\multicolumn{1}{c}{52.3$/$65.0}&\multicolumn{1}{|c}{71.7$/$95.2}&\multicolumn{1}{c}{42.0$/$91.8}&\multicolumn{1}{c}{92.8$/$48.1}&\multicolumn{1}{c}{92.8$/$52.0}&\multicolumn{1}{c}{74.8$/$71.8}\\
Ummkd~\cite{dou2020unpaired}&\multicolumn{1}{c|}{50}&\multicolumn{1}{c}{83.6$/$39.6}&\multicolumn{1}{c}{36.9$/$\textcolor{blue}{\textbf{90.7}}}&\multicolumn{1}{c}{\textcolor{blue}{\textbf{93.0}}$/$\textcolor{blue}{\textbf{91.1}}}&\multicolumn{1}{c}{\textcolor{blue}{\textbf{60.1}}$/$\textcolor{red}{\textbf{87.8}}}&\multicolumn{1}{c}{\textcolor{blue}{\textbf{68.4}}$/$77.3}&\multicolumn{1}{|c}{65.4$/$\textcolor{blue}{\textbf{95.3}}}&\multicolumn{1}{c}{72.2$/$92.0}&\multicolumn{1}{c}{92.9$/$61.1}&\multicolumn{1}{c}{92.3$/$25.8}&\multicolumn{1}{c}{80.7$/$68.5}\\
\hline
AdaOutput~\cite{tsai2018learning}&\multicolumn{1}{c|}{50}&\multicolumn{1}{c}{81.1$/$83.4}&\multicolumn{1}{c}{24.8$/$88.4}&\multicolumn{1}{c}{85.9$/$85.5}&\multicolumn{1}{c}{50.2$/$85.7}&\multicolumn{1}{c}{60.5$/$85.8}&\multicolumn{1}{|c}{77.1$/$94.0}&\multicolumn{1}{c}{25.3$/$91.8}&\multicolumn{1}{c}{\textcolor{blue}{\textbf{93.8}}$/$78.6}&\multicolumn{1}{c}{\textcolor{blue}{\textbf{93.5}}$/$38.1}&\multicolumn{1}{c}{72.4$/$75.6}\\
SIFA-v2~\cite{chen2020unsupervised}&\multicolumn{1}{c|}{50}&\multicolumn{1}{c}{82.7$/$\textcolor{blue}{\textbf{84.6}}}&\multicolumn{1}{c}{35.1$/$87.4}&\multicolumn{1}{c}{92.3$/$89.2}&\multicolumn{1}{c}{58.3$/$85.4}&\multicolumn{1}{c}{67.1$/$\textcolor{blue}{\textbf{86.6}}}&\multicolumn{1}{|c}{76.4$/$92.1}&\multicolumn{1}{c}{\textcolor{red}{\textbf{90.5}}$/$88.6}&\multicolumn{1}{c}{90.0$/$\textcolor{blue}{\textbf{86.3}}}&\multicolumn{1}{c}{90.8$/$\textcolor{blue}{\textbf{71.3}}}&\multicolumn{1}{c}{\textcolor{blue}{\textbf{86.9}}$/$\textcolor{blue}{\textbf{84.6}}} \\
ProCA~\cite{jiang2022prototypical}&\multicolumn{1}{c|}{50}&\multicolumn{1}{c}{82.6$/$79.0}&\multicolumn{1}{c}{\textcolor{blue}{\textbf{54.7}}$/$87.7}&\multicolumn{1}{c}{86.9$/$87.7}&\multicolumn{1}{c}{41.2$/$83.5}&\multicolumn{1}{c}{66.4$/$84.5}&\multicolumn{1}{|c}{74.6$/$95.0}&\multicolumn{1}{c}{68.2$/$\textcolor{red}{\textbf{93.0}}}&\multicolumn{1}{c}{91.5$/$78.0}&\multicolumn{1}{c}{93.3$/$32.2}&\multicolumn{1}{c}{81.9$/$74.5} \\
ProDA~\cite{zhang2021prototypical}&\multicolumn{1}{c|}{50}&\multicolumn{1}{c}{\textcolor{blue}{\textbf{83.7}}$/$73.3}&\multicolumn{1}{c}{44.4$/$90.2}&\multicolumn{1}{c}{85.3$/$89.2}&\multicolumn{1}{c}{47.5$/$85.9}&\multicolumn{1}{c}{65.2$/$84.7}&\multicolumn{1}{|c}{\textcolor{blue}{\textbf{80.7}}$/$94.6}&\multicolumn{1}{c}{84.7$/$92.0}&\multicolumn{1}{c}{88.9$/$75.9}&\multicolumn{1}{c}{91.6$/$60.6}&\multicolumn{1}{c}{86.5$/$80.7} \\
\hline
\textbf{Ours (DEST)}&\multicolumn{1}{c|}{50}&\multicolumn{1}{c}{\textcolor{red}{\textbf{85.5}}$/$\textcolor{red}{\textbf{88.8}}}&\multicolumn{1}{c}{\textcolor{red}{\textbf{82.0}}$/$\textcolor{red}{\textbf{91.2}}}&\multicolumn{1}{c}{\textcolor{red}{\textbf{93.3}}$/$\textcolor{red}{\textbf{92.9}}}&\multicolumn{1}{c}{\textcolor{red}{\textbf{66.7}}$/$\textcolor{blue}{\textbf{87.2}}}&\multicolumn{1}{c}{\textcolor{red}{\textbf{81.9}}$/$\textcolor{red}{\textbf{90.0}}}&\multicolumn{1}{|c}{\textcolor{red}{\textbf{89.1}}$/$\textcolor{red}{\textbf{95.6}}}&\multicolumn{1}{c}{\textcolor{blue}{\textbf{90.2}}$/$\textcolor{blue}{\textbf{92.7}}}&\multicolumn{1}{c}{\textcolor{red}{\textbf{93.9}}$/$\textcolor{red}{\textbf{86.9}}}&\multicolumn{1}{c}{\textcolor{red}{\textbf{93.6}}$/$\textcolor{red}{\textbf{87.4}}}&\multicolumn{1}{c}{\textcolor{red}{\textbf{91.7}}$/$\textcolor{red}{\textbf{90.7}}} \\
\Xhline{4\arrayrulewidth}
Ummkd~\cite{dou2020unpaired} (full)&\multicolumn{1}{c|}{100}&\multicolumn{1}{c}{82.8$/$93.3}&\multicolumn{1}{c}{85.7$/$91.6}&\multicolumn{1}{c}{93.4$/$92.0}&\multicolumn{1}{c}{79.8$/$87.6}&\multicolumn{1}{c}{85.4$/$91.1}&\multicolumn{1}{|c}{95.0$/$95.8}&\multicolumn{1}{c}{95.1$/$92.8}&\multicolumn{1}{c}{94.7$/$92.4}&\multicolumn{1}{c}{92.8$/$94.1}&\multicolumn{1}{c}{94.4$/$93.8} \\
\hline
\end{tabular}}
\end{center}
\vspace{-8mm}
\label{table4}
\end{table}

\noindent\textbf{Quantitative Comparison.} We compare our framework with state-of-the-art partially supervised segmentation methods: ConCNN~\cite{dmitriev2019learning}, CoWM~\cite{huang2020multi}, and Marginal~\cite{shi2021marginal}; unpaired multi-modal learning methods: Y-Network~\cite{valindria2018multi}, X-network \cite{valindria2018multi}, and Ummkd~\cite{dou2020unpaired}; and unsupervised domain adaptation methods which reduce the distribution discrepancy across modalities: AdaOutput~\cite{tsai2018learning}, SIFA-v2~\cite{chen2020unsupervised}, ProCA~\cite{jiang2022prototypical}, and ProDA~\cite{zhang2021prototypical}. We implement a "Supervised (partial)" method, which trains two segmentation networks with partial labels in each modality as a lower bound. We implement a "Ummkd (full)" method, which trains the state-of-the-art Ummkd method for unpaired multi-modal learning with full annotations as an upper bound. 
Table~\ref{table4} presents the quantitative comparison results for scenario (1) tasks. We have the following observations: \textbf{(a).} partially supervised segmentation methods generally performs better than the baseline method, as they can leverage all partially labeled datasets to train a segmentation network, however they fall short in exploiting the multi-modal information in PSUMML; \textbf{(b).} unpaired multi-modal learning methods can leverage partially labeled unpaired multi-modal data for learning. However, due to partial class distribution discrepancy, they perform worse on partially labeled classes in each modality; \textbf{(c).} unsupervised domain adaptation methods reduce the overall distribution discrepancy across modalities. We observe distribution discrepancy reduction effect with improved partially labeled class performance, but they perform worse than our principally designed decomposed partial class adaptation loss; \textbf{(d).} our framework outperforms existing methods significantly in two segmentation tasks. Specifically, our framework outperforms the second best by \textbf{\plus13.5} and \textbf{\plus3.4} points in dice score 
for MRI and CT images in cardiac substructure segmentation and outperforms the second best by \textbf{\plus4.8} points and \textbf{\plus6.1} points in dice score 
for MRI and CT images in multi-organ segmentation. With 50\% of annotation, the performance of our framework approaches state-of-the-art Ummkd method with full annotation for both tasks. (Please view the Appendix for \textbf{ASD experiment results} and \textbf{visualization comparison results}.)

\begin{table}[t]
    \begin{minipage}[t]{.46\linewidth}
    \caption{Performance comparison on cardiac substructure segmentation under PSUMML scenario (2) and (3) task.}
\begin{center}
\resizebox{0.95\linewidth}{!}{%
\begin{tabular}{c|p{0.045\linewidth}p{0.045\linewidth}p{0.045\linewidth}p{0.045\linewidth}p{0.045\linewidth}p{0.045\linewidth}}
\hline
&\multicolumn{3}{c}{Scenario (2)} & \multicolumn{3}{|c}{Scenario (3)} \\
\cline{2-7}
\multicolumn{1}{c|}{Method}&\multicolumn{1}{c|}{Anno.}&\multicolumn{2}{c}{Dice}&\multicolumn{1}{|c|}{Anno.}&\multicolumn{2}{c}{Dice} \\
\cline{3-4}\cline{6-7}
&\multicolumn{1}{c|}{Cost (\%)}&\multicolumn{1}{c|}{MRI}&\multicolumn{1}{c}{CT}&\multicolumn{1}{|c|}{Cost (\%)}&\multicolumn{1}{c|}{MRI}&\multicolumn{1}{c}{CT}\\
\hline
Supervised (partial)&\multicolumn{1}{c|}{75}&\multicolumn{1}{c|}{65.5}&\multicolumn{1}{c}{68.6}&\multicolumn{1}{|c|}{75}&\multicolumn{1}{c|}{85.2}&\multicolumn{1}{c}{49.1} \\
\Xhline{3\arrayrulewidth}
Marginal~\cite{shi2021marginal}&\multicolumn{1}{c|}{75}&\multicolumn{1}{c|}{66.0}&\multicolumn{1}{c}{69.6}&\multicolumn{1}{|c|}{75}&\multicolumn{1}{c|}{85.0}&\multicolumn{1}{c}{84.4} \\
Ummkd~\cite{dou2020unpaired}&\multicolumn{1}{c|}{75}&\multicolumn{1}{c|}{81.6}&\multicolumn{1}{c}{84.6}&\multicolumn{1}{|c|}{75}&\multicolumn{1}{c|}{84.9}&\multicolumn{1}{c}{85.9} \\
SIFA-v2~\cite{chen2020unsupervised}&\multicolumn{1}{c|}{75}&\multicolumn{1}{c|}{74.7}&\multicolumn{1}{c}{86.1}&\multicolumn{1}{|c|}{75}&\multicolumn{1}{c|}{85.2}&\multicolumn{1}{c}{85.0} \\
\hline
\textbf{Ours (DEST)}&\multicolumn{1}{c|}{75}&\multicolumn{1}{c|}{\textbf{82.1}}&\multicolumn{1}{c}{\textbf{90.1}}&\multicolumn{1}{|c|}{75}&\multicolumn{1}{c|}{\textbf{86.1}}&\multicolumn{1}{c}{\textbf{89.6}}\\
\Xhline{3\arrayrulewidth}
Ummkd~\cite{dou2020unpaired} (full)&\multicolumn{1}{c|}{100}&\multicolumn{1}{c|}{85.4}&\multicolumn{1}{c}{91.1}&\multicolumn{1}{|c|}{100}&\multicolumn{1}{c|}{85.4}&\multicolumn{1}{c}{91.1}\\
\hline
\end{tabular}}
\end{center}
\label{plcpa_table3}
    \end{minipage}%
    \hfill
    \begin{minipage}[t]{.45\linewidth}
        \begin{center}
\caption{Ablation Study on cardiac substructure segmentation under PSUMML scenario (1) task.}
\label{table3}
\resizebox{0.95\linewidth}{!}{%
\begin{tabular}{c|cc|c|c|c|c}
\hline
\multirow{2}{*}{Method}&\multirow{2}{*}{$\mathcal{L}_{PCL}$}&\multirow{2}{*}{$\mathcal{L}_{DPCA}$}&\multicolumn{2}{c|}{$\mathcal{L}_{SEST}$}&\multicolumn{2}{c}{Dice}\\
\cline{4-7}
&&&\multirow{1}{*}{W/o Mod.}&\multirow{1}{*}{Mod.}&\multirow{1}{*}{MRI}&\multirow{1}{*}{CT}\\
\hline
PCL&\cmark&\xmark&\xmark&\xmark&65.1&83.7 \\
\hline
+DPCA&\cmark&\cmark&\xmark&\xmark&79.9&88.2\\
\hline
DEST [K=4] W/o Mod.&\cmark&\cmark&\cmark&\xmark&80.9&89.2\\
\hline
DEST [K=1]&\cmark&\cmark&\xmark&\cmark&81.0&88.3\\
\hline
DEST [K=2]&\cmark&\cmark&\xmark&\cmark&81.0&89.8\\
\hline
DEST [K=4]&\cmark&\cmark&\xmark&\cmark&\textbf{81.9}&90.0\\
\hline
DEST [K=8]&\cmark&\cmark&\xmark&\cmark&81.5&\textbf{90.3}\\
\hline
\end{tabular}}
\end{center}
    \end{minipage} 
\vspace{-8mm}
\end{table}

\noindent\textbf{Performance Under Different Scenarios.} We further investigate the other two scenarios of PSUMML for cardiac substructure segmentation: for scenario (2), we let MRI data be partially labeled with AA, LAC, and LVC and CT data be partially labeled with LAC, LVC, and MYO; for scenario (3), we let MRI data be fully labeled and CT data be partially labeled with LAC and MYO. Table~\ref{plcpa_table3} shows the experiment results. Our method outperforms existing methods significantly and approaches Ummkd (full) method with 25\% cost reduction.

\noindent\textbf{Ablation Study.} Table~\ref{table3} shows the ablation study results.
Plain partial class learning does not give satisfying performance. Decomposed partial class adaptation helps to improve the performance by a large margin. Snapshot ensembled self-training further improves the model segmentation performance. We observe noise modulation is beneficial and when we increase the snapshot ensemble size $k$, the segmentation performance gradually improves and plateaus at $k=4$.

\section{Conclusion}
In this paper, we introduce partially supervised unpaired multi-modal learning for label-efficient medical image segmentation. We theoretically analyze the new machine learning problem and propose a novel decomposed partial class adaptation with snapshot ensembled self-training framework which significantly outperforms existing methods in two segmentation tasks. Our proposal serves to solve the time-consuming and costly annotation problem in medical domain.

\noindent \textbf{Acknowledgement}
This work was supported by the Agency for Science, Technology, and Research (A*STAR) through its AME Programmatic Funding Scheme Under Project A20H4b0141, the National Research Foundation (NRF) Singapore under its AI Singapore Programme (AISG Award No: AISG2-TC-2021-003), the Agency for Science, Technology, and Research (A*STAR) through its RIE2020 Health and Biomedical Sciences (HBMS) Industry Alignment Fund Pre-Positioning (IAF-PP) (grant no. H20C6a0032), the 2022 Horizontal Technology Coordinating Office Seed Fund (Biomedical Engineering Programme – BEP RUN 3, grant no. C221318005) and partially supported by A*STAR Central Research Fund "A Secure and Privacy-Preserving AI Platform for Digital Health”.

\bibliographystyle{splncs04}
\bibliography{main}

\begin{thebibliography}{10}
\providecommand{\url}[1]{\texttt{#1}}
\providecommand{\urlprefix}{URL }
\providecommand{\doi}[1]{https://doi.org/#1}

\bibitem{chen2020unsupervised}
Chen, C., Dou, Q., Chen, H., Qin, J., Heng, P.A.: Unsupervised bidirectional cross-modality adaptation via deeply synergistic image and feature alignment for medical image segmentation. IEEE Transactions on Medical Imaging  (2020)

\bibitem{dmitriev2019learning}
Dmitriev, K., Kaufman, A.E.: Learning multi-class segmentations from single-class datasets. In: Proceedings of the IEEE/CVF Conference on Computer Vision and Pattern Recognition. pp. 9501--9511 (2019)

\bibitem{dou2020unpaired}
Dou, Q., Liu, Q., Heng, P.A., Glocker, B.: Unpaired multi-modal segmentation via knowledge distillation. IEEE transactions on medical imaging  \textbf{39}(7),  2415--2425 (2020)

\bibitem{goodfellow2014generative}
Goodfellow, I., Pouget-Abadie, J., Mirza, M., Xu, B., Warde-Farley, D., Ozair, S., Courville, A., Bengio, Y.: Generative adversarial nets. In: Advances in neural information processing systems. pp. 2672--2680 (2014)

\bibitem{huang2017snapshot}
Huang, G., Li, Y., Pleiss, G., Liu, Z., Hopcroft, J.E., Weinberger, K.Q.: Snapshot ensembles: Train 1, get m for free. arXiv preprint arXiv:1704.00109  (2017)

\bibitem{huang2020multi}
Huang, R., Zheng, Y., Hu, Z., Zhang, S., Li, H.: Multi-organ segmentation via co-training weight-averaged models from few-organ datasets. In: International Conference on Medical Image Computing and Computer-Assisted Intervention. pp. 146--155. Springer (2020)

\bibitem{Ioffe2015BatchNA}
Ioffe, S., Szegedy, C.: Batch normalization: Accelerating deep network training by reducing internal covariate shift. ArXiv  \textbf{abs/1502.03167} (2015)

\bibitem{jiang2022prototypical}
Jiang, Z., Li, Y., Yang, C., Gao, P., Wang, Y., Tai, Y., Wang, C.: Prototypical contrast adaptation for domain adaptive semantic segmentation. In: Computer Vision--ECCV 2022: 17th European Conference, Tel Aviv, Israel, October 23--27, 2022, Proceedings, Part XXXIV. pp. 36--54. Springer (2022)

\bibitem{kavur2020chaos}
Kavur, A.E., Gezer, N.S., Bar{\i}{\c{s}}, M., Conze, P.H., Groza, V., Pham, D.D., Chatterjee, S., Ernst, P., {\"O}zkan, S., Baydar, B., et~al.: Chaos challenge--combined (ct-mr) healthy abdominal organ segmentation. arXiv preprint arXiv:2001.06535  (2020)

\bibitem{landman2017}
Landman, B., Xu, Z., Iglesias, J.E., Styner, M., Langerak, T.R., Klein, A.: Multi-atlas labeling beyond the cranial vault  (2020)

\bibitem{milletari2016v}
Milletari, F., Navab, N., Ahmadi, S.A.: V-net: Fully convolutional neural networks for volumetric medical image segmentation. In: 2016 fourth international conference on 3D vision (3DV). pp. 565--571. IEEE (2016)

\bibitem{shi2021marginal}
Shi, G., Xiao, L., Chen, Y., Zhou, S.K.: Marginal loss and exclusion loss for partially supervised multi-organ segmentation. Medical Image Analysis  \textbf{70},  101979 (2021)

\bibitem{tsai2018learning}
Tsai, Y.H., Hung, W.C., Schulter, S., Sohn, K., Yang, M.H., Chandraker, M.: Learning to adapt structured output space for semantic segmentation. In: Proceedings of the IEEE Conference on Computer Vision and Pattern Recognition. pp. 7472--7481 (2018)

\bibitem{valindria2018multi}
Valindria, V.V., Pawlowski, N., Rajchl, M., Lavdas, I., Aboagye, E.O., Rockall, A.G., Rueckert, D., Glocker, B.: Multi-modal learning from unpaired images: Application to multi-organ segmentation in ct and mri. In: 2018 IEEE winter conference on applications of computer vision (WACV). pp. 547--556. IEEE (2018)

\bibitem{yang2023toward}
Yang, J., Zhu, Y., Wang, C., Li, Z., Zhang, R.: Toward unpaired multi-modal medical image segmentation via learning structured semantic consistency. In: Medical Imaging with Deep Learning (2023), \url{https://openreview.net/forum?id=e9qGhrfP1v}

\bibitem{zhang2022deep}
Zhang, K., Zhuang, X.: Deep compatible learning for partially-supervised medical image segmentation. arXiv preprint arXiv:2206.09148  (2022)

\bibitem{zhang2021prototypical}
Zhang, P., Zhang, B., Zhang, T., Chen, D., Wang, Y., Wen, F.: Prototypical pseudo label denoising and target structure learning for domain adaptive semantic segmentation. In: Proceedings of the IEEE/CVF conference on computer vision and pattern recognition. pp. 12414--12424 (2021)

\bibitem{zhou2012ensemble}
Zhou, Z.H.: Ensemble methods: foundations and algorithms. CRC press (2012)

\bibitem{zhuang2016multi}
Zhuang, X., Shen, J.: Multi-scale patch and multi-modality atlases for whole heart segmentation of mri. Medical image analysis  \textbf{31},  77--87 (2016)

\end{thebibliography}

\end{document}